\documentclass{article}

\usepackage[preprint]{neurips_2026}

\usepackage[utf8]{inputenc}
\usepackage[T1]{fontenc}
\usepackage[table]{xcolor}

\usepackage{url}
\usepackage{graphicx}
\usepackage{booktabs}
\usepackage{longtable}
\usepackage{float}
\usepackage{multirow}
\usepackage{threeparttable}
\usepackage{makecell}
\usepackage{amsmath}
\usepackage{amssymb}
\usepackage{amsfonts}
\usepackage{nicefrac}
\usepackage{microtype}
\usepackage{listings}
\usepackage[most]{tcolorbox}
\usepackage{pifont}
\usepackage{inconsolata}
\usepackage{capt-of}
\usepackage{hyperref}
\definecolor{codebg}{RGB}{245,245,245}
\definecolor{codeframe}{RGB}{160,160,160}
\definecolor{codecomment}{RGB}{0,140,0}
\definecolor{codekeyword}{RGB}{0,0,180}
\definecolor{codestring}{RGB}{170,40,40}

\lstdefinestyle{mypython}{
  language=Python,
  basicstyle=\ttfamily\footnotesize,
  keywordstyle=\color{codekeyword},
  commentstyle=\color{codecomment},
  stringstyle=\color{codestring},
  showstringspaces=false,
  breaklines=true,
  breakatwhitespace=false,
  columns=fullflexible,
  keepspaces=true,
  tabsize=2
}

\newtcblisting{pycode}{
  listing engine=listings,
  listing only,
  colback=codebg,
  colframe=codeframe,
  boxrule=0.4pt,
  arc=0pt,
  outer arc=0pt,
  left=0.4em,
  right=0.4em,
  top=0.35em,
  bottom=0.35em,
  boxsep=0pt,
  listing options={style=mypython}
}

\newcommand{\na}{--}

\newcommand{\cmark}{\textcolor{green!50!black}{\ding{51}}}
\newcommand{\xmark}{\textcolor{red!70!black}{\ding{55}}}

\title{TIDE-Bench: Task-Aware and Diagnostic Evaluation of Tool-Integrated Reasoning}

\author{%
Yize Li\thanks{Equal contribution.} \\
University of Chinese Academy of Sciences \\ Institute of Software,\\Chinese Academy of Sciences\\ 
\texttt{liyize251@mails.ucas.ac.cn}
\And
Junzhi Li\footnotemark[1]  \\
University of Chinese Academy of Sciences \\ Institute of Software,\\ Chinese Academy of Sciences\\ 
\texttt{lijunzhi25@mails.ucas.ac.cn}
\And
Jason Song\footnotemark[1] \\
University of Chinese Academy of Sciences \\ Institute of Software, \\Chinese Academy of Sciences\\ 
\texttt{purpnm2002@gmail.com}
\And
Chuxiong Sun \\
Institute of Software, \\Chinese Academy of Sciences\\ 
\texttt{chuxiong2016@iscas.ac.cn}
\And
Rui Wang \\
Institute of Software, \\Chinese Academy of Sciences\\ 
\texttt{wangrui@iscas.ac.cn}
\And
Changwen Zheng\thanks{Corresponding author.}  \\
Institute of Software, \\Chinese Academy of Sciences\\ 
\texttt{changwen@iscas.ac.cn}
}

\begin{document}
\maketitle
\begin{abstract}
Tool-integrated reasoning has emerged as a promising paradigm for enhancing large language models with external computation, retrieval, and execution capabilities. However, the field still lacks a high-quality and unified evaluation benchmark, and existing TIR evaluations remain limited in dataset quality, task diversity, diagnostic comprehensiveness, and evaluation efficiency. In this work, we introduce TIDE-Bench, a holistic and efficient benchmark for evaluating TIR methods, featuring three key advantages. First, it provides diverse task settings, combining widely used mathematical reasoning and knowledge-intensive QA tasks with two newly designed tasks—tool-grounded experimental design task and dynamic interactive task—to probe models’ abilities in complex tool invocation and multi-tool coordination. Second, TIDE-Bench adopts a comprehensive yet task-aware evaluation protocol, jointly measuring final answer quality, process reliability, tool-use efficiency, and inference cost across heterogeneous task settings. Third, TIDE-Bench constructs high-quality and discriminative evaluation sets by filtering low-discrimination instances from existing datasets, substantially reducing evaluation cost while focusing on more challenging samples. Extensive experiments on multiple foundation models and TIR methods reveal persistent bottlenecks in tool grounding, offering insights for future TIR research.
\end{abstract}

\section{Introduction}

\begin{figure*}[t]
    \centering
    \includegraphics[width=\textwidth]{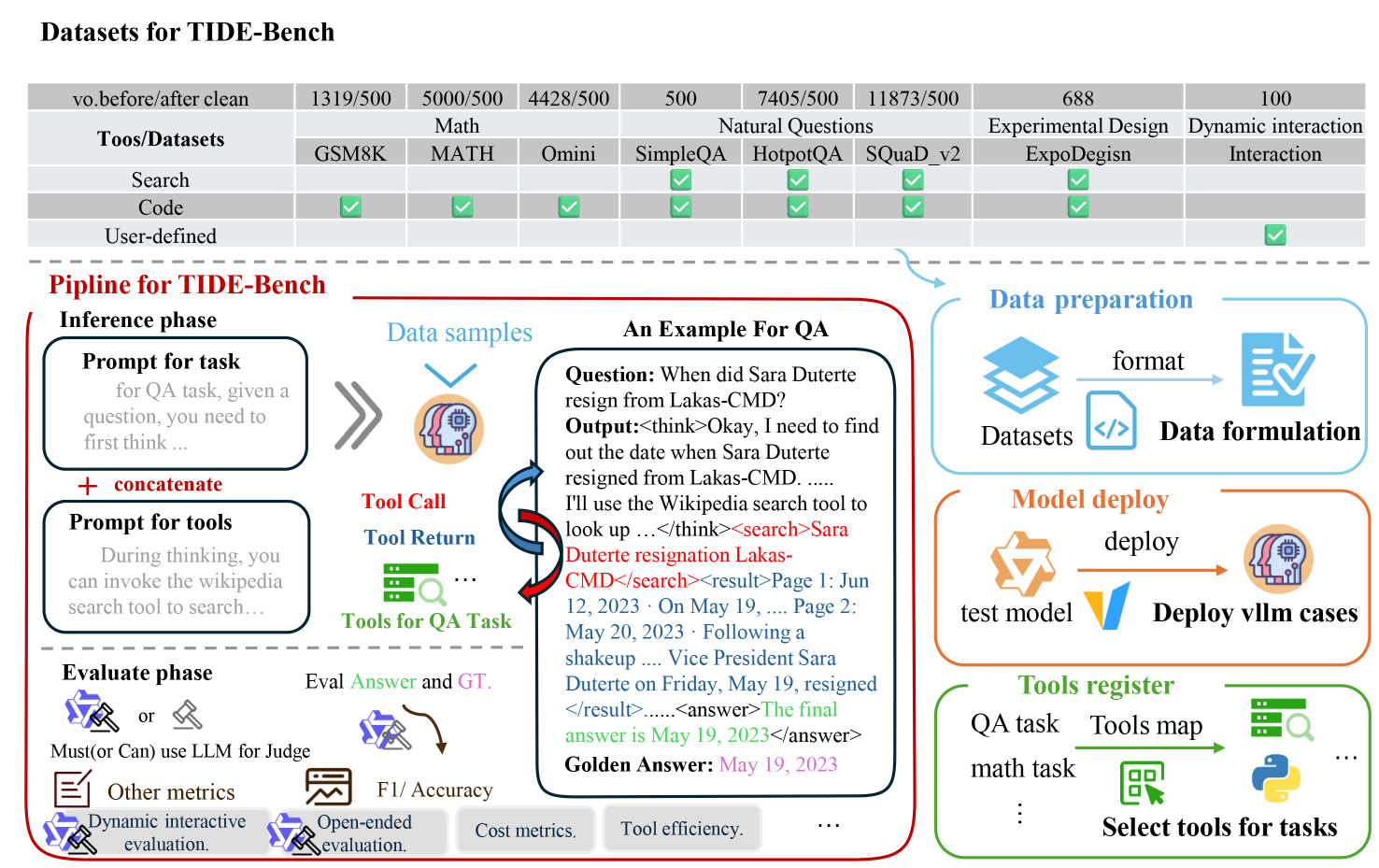}
    \caption{The overall framework of TIDE-Bench. The upper part summarizes the benchmark datasets and the tool types supported by each task. And the lower part illustrates the inference and evaluation pipeline of TIDE-Bench.More details about user-defined tools in Interaction is provided in Appendix~\ref{app:user_defined_tools}.}
    \label{fig:TIDE-Bench-framework}
\end{figure*}

Large Language Models (LLMs) have achieved significant progress in complex reasoning tasks across diverse domains. 
However, reasoning solely from internal model parameters remains insufficient for many real-world problems: model knowledge can be outdated or incomplete, symbolic operations require precise executable computation, and dynamic tasks often demand interaction with external environments or feedback signals. 
Tool-Integrated Reasoning (TIR) has therefore emerged as a promising paradigm that augments LLM reasoning with external retrieval, computation, code execution, and environment interaction \citep{yao2023react,schick2023toolformer,qin2023toolllm,zhao2025dissecting}. 
As tool use becomes central to agentic systems, it is increasingly important to evaluate whether models can invoke appropriate tools, interpret their outputs, and ground final responses in tool-mediated evidence.

Despite recent progress, existing TIR evaluations remain limited in the following three aspects. First, \textbf{existing evaluation tasks lack sufficient diversity and quality}. In the absence of a unified benchmark for TIR, prior methods typically rely on ad hoc combinations of standard math reasoning and knowledge-intensive QA datasets, such as GSM8K and MATH for code-assisted mathematical reasoning \citep{cobbe2021gsm8k,hendrycks2021math}, and HotpotQA and MuSiQue for retrieval-augmented or multi-hop question answering \citep{yang2018hotpotqa,trivedi2022musique}.
However, these datasets instantiate TIR mainly as tool use over static instances, such as code execution for math or retrieval for closed-ended QA. As a result, these tasks provide limited support for evaluating complex tool-use capabilities, including the ability to coordinate heterogeneous tools in open-ended problem solving, translate underspecified goals into executable tool-grounded procedures, and maintain reliable tool use across multi-turn and stateful interactions\citep{lu2025toolsandbox,yao2024tau}.

Second, existing evaluations often \textbf{lack diagnostic granularity} to measure how tools contribute to reasoning. Most benchmarks only report answer-level scores such as accuracy or pass@k, while only some additionally report the number or success rate of tool calls. These metrics do not show whether a correct answer is grounded in tool outputs, whether the model misuses retrieved evidence or execution results, or whether success relies on many unnecessary calls and long contexts. 

Third, existing evaluations also \textbf{lack sufficient efficiency and scalability} to support iterative model development. As LLMs continue to improve, early benchmarks become less discriminative, with many instances now solvable by foundation models even without invoking external tools. These low-discrimination instances provide limited value for evaluating tool-integrated reasoning, yet they still incur substantial inference time, tool-calling overhead, and long-context computation when included in large evaluation suites. This makes iterative comparison of models and tool-use strategies unnecessarily costly \citep{chen2026agentcpm,team2025minicpm4}.
To address these limitations, we introduce the \textbf{T}ool-\textbf{I}ntegrated \textbf{D}iagnostic \textbf{E}valuation \textbf{Bench}mark (TIDE-Bench), a diverse, diagnostic, and efficient benchmark for TIR. As illustrated in Figure~\ref{fig:TIDE-Bench-framework}, TIDE-Bench combines community-standard mathematical and knowledge-intensive QA datasets with two new tracks: Tool-Grounded Experimental Design (ExpoDesign) and Dynamic Interaction (Interaction).
This design preserves comparability with established closed-ended benchmarks while extending evaluation to discovery-oriented experimental planning and deployment-oriented interactive tool orchestration that requires complex tool invocation and multi-tool coordination.
TIDE-Bench also adopts a unified task-aware protocol to evaluate answer quality, process reliability, tool-use efficiency, latency, and token cost, diagnosing whether tools improve reasoning reliably and economically.
Finally, it filters low-discrimination samples from existing datasets, reducing benchmark cost while retaining representative task difficulty.

Our main contributions are:

1. We introduce TIDE-Bench, a unified benchmark for TIR that preserves comparability with standard math and knowledge-intensive QA tasks while extending evaluation to two underexplored settings: ExpoDesign and Interaction.

2. We construct a compact yet discriminative evaluation suite through systematic sample filtering. By removing redundant and low-discrimination instances with a non-tool screening model, TIDE-Bench reduces large QA subsets to 500 representative examples each, saving more than 6 hours in a single evaluation while retaining task difficulty.

3. We propose a task-aware evaluation protocol that measures final outcomes, reasoning processes, and tool-use cost. Beyond answer quality, TIDE-Bench diagnoses whether tool calls are grounded in the reasoning process, contribute to reliable answers, and justify their computational cost.

4. We conduct systematic experiments on multiple open-source and proprietary models, comparing tool-enabled and no-tool settings where applicable. The results show that tool benefits are strongly task-dependent: retrieval-heavy tasks benefit more consistently, while the benchmark exposes persistent challenges in multi-hop evidence composition, faithful mathematical formalization, and multi-turn tool orchestration.

\section{Related Work}
\textbf{Tool-Use Evaluation.}
Existing evaluations of tool use have primarily focused on whether models can select appropriate tools, generate valid API or function calls, and execute them successfully. Representative benchmarks construct executable API environments, scale evaluation to large collections of real-world tools, or assess structured serial and parallel function calling under predefined schemas \citep{li2023apibank,patil2024gorilla,qin2023toolllm,patil2025berkeley}. Subsequent studies further decompose, consolidate, or extend these evaluations by examining benchmark design, tool-learning protocols, and models' decisions about when and what types of tools to invoke \citep{chen2024t,chen2025toleap,huang2024metatool,ning2024wtueval}. While these studies provide an important foundation for evaluating tool-augmented language models, their focus remains largely on tool-use behavior itself. In contrast, TIR evaluates whether models can interpret tool outputs, integrate them into intermediate reasoning, and use them to produce reliable final answers. Thus, tool-selection accuracy, invocation success, or execution correctness alone cannot fully characterize TIR capabilities.

\textbf{TIR Evaluation Practices. }
Recent TIR studies have developed diverse methods for improving tool-integrated reasoning, including reasoning-trajectory construction, automated tool invocation, reinforcement learning, search-augmented reasoning, and tool-preference optimization \citep{xue2025simpletir,wei2025autotir,qian2025toolrl,chen2025toward,jin2025search,dong2025tool,zhang2025criticsearch}. Their evaluations commonly rely on mathematical reasoning, retrieval-augmented question answering, demonstrating the utility of tools such as search and code interpreters for complex reasoning. However, these evaluations are primarily method-oriented: they are designed to measure performance gains on specific task settings, with metrics such as accuracy, F1, pass@k, or tool-call counts. As a result, they offer limited diagnostic insight into when tools help, whether tool outputs are reliably grounded in reasoning, and how gains trade off against cost across task formats.

\textbf{Generalist Agent Evaluation. }
A related line of work evaluates generalist agents in interactive or real-world environments. Representative benchmarks assess multi-step decision making, environment interaction, web navigation, computer use, and broader abilities involving reasoning, tool use, multimodal understanding, and information integration \citep{liu2023agentbench,mialon2023gaia,zhou2023webarena,deng2023mind2web,xie2024osworld}. While these benchmarks are closer to practical applications, they evaluate broad agent performance and often entangle planning, exploration, interaction, perception, tool invocation, and long-horizon task completion. This makes it difficult to isolate TIR-specific bottlenecks; we instead target whether models can invoke tools across task formats and turn tool outputs into accurate, cost-effective reasoning.

\begin{table*}[t]
\caption{Performance of Qwen3-4b before and after cleaning on Natural Questions and MATH benchmarks.Among them, for the SimpleQA dataset, we directly used the dataset provided in Tool-Light\citep{chen2025toward} rather than the original source dataset, which contains 4,326 samples.\citep{wei2024simpleqa}}
\label{tab:cleaning_results}
\centering
\small
\setlength{\tabcolsep}{6pt}
\renewcommand{\arraystretch}{1.2}
\resizebox{\textwidth}{!}{%
\begin{tabular}{llcccccc}
\toprule
\multirow{2}{*}{} & \multirow{2}{*}{}
& \multicolumn{3}{c}{Natural Questions}
& \multicolumn{3}{c}{Math} \\
\cmidrule(lr){3-5} \cmidrule(lr){6-8}
& & SimpleQA & HotpotQA & SQuAD\_v2& GSM8K & MATH & OmniMath\\
\midrule
\multirow{3}{*}{Before Cleaning}
& samples           & 500   & 7405   & 11873  & 1319 & 500 & 525 \\
& test time (s)     & 286.2 & 7383.7 & 8839.8 & 1030 & 859 & 918 \\
& f1 score/accuracy & 0.08  & 0.44   & 0.65   & 0.81 & 0.49 & 0.12 \\
\midrule
\multirow{3}{*}{After Cleaning}
& samples           & 500   & 500    & 500    & 500  & 500 & 500 \\
& test time (s)     & 286.2   & 498.6  & 372.3  & 386.7 & 859 & 878 \\
& f1 score/accuracy & 0.08   & 0      & 0      & 0.55 & 0.49 & 0.09 \\
\bottomrule
\end{tabular}
}
\end{table*}
\section{Method}


\subsection{High-Quality and Diverse Dataset Reconstruction}



In the absence of a unified benchmark specifically designed for tool-integrated reasoning, existing TIR methods typically construct their evaluations from ad hoc combinations of standard QA and mathematical reasoning datasets. We analyze these commonly used evaluation sources and identify two limitations that hinder a reliable and comprehensive assessment of TIR abilities.

First, existing QA and mathematical reasoning datasets contain many low-discrimination samples that can be solved effectively without tool use. On popular datasets such as HotpotQA \citep{yang2018hotpotqa}, SQuADv2 \citep{rajpurkar2018know}, and GSM8K \citep{cobbe2021gsm8k}, we find that even a relatively small model such as Qwen3-4B can solve a substantial portion of samples without invoking external tools. These samples provide limited value for TIR evaluation, as they make it difficult to isolate the contribution of tool use from the model's internal reasoning ability. We therefore filter out such low-discrimination samples by evaluating Qwen3-4B in a tool-free setting and retaining the 500 most challenging samples from each dataset. This improves dataset quality and evaluation efficiency by focusing the benchmark on samples that better reflect the need for tool-integrated reasoning. Table~\ref{tab:cleaning_results} reports the evaluation and cleaning results, and Appendix~\ref{sec:appendix} details the corresponding data-cleaning procedure, screening configuration, and evaluation hyperparameters.

\begin{table*}[!t]
\centering
\small
\caption[Coverage of evaluation tasks in TIR-related lines]{Coverage of reported evaluation suites in representative TIR lines. \textbf{NQ}/Search: retrieval- or web-search-based QA; \textbf{Math}/Code: math with code execution; \textbf{Interaction} and \textbf{ExpoDesign} denote the two newly introduced tracks.}

\label{tab:benchmarks}

\renewcommand{\arraystretch}{1.12}

\begin{tabular}{lcccc}
\toprule
\multirow{2}{*}{\textbf{Benchmark}} & \textbf{Natural Questions} & \textbf{Math} & \textbf{Interaction} & \textbf{ExpoDesign} \\
\cmidrule(lr){2-2} \cmidrule(lr){3-3} \cmidrule(lr){4-4} \cmidrule(lr){5-5} 
& Search & Code & User-defined & Code+Search \\
\midrule

\rowcolor{gray!6}
\textbf{TIDE-Bench} (ours) & \cmark & \cmark & \cmark & \cmark \\
Understand TIR\citep{lin2025understanding} & \xmark & \cmark & \xmark & \xmark \\
\rowcolor{gray!6}
Tool-Light\citep{chen2025toward} & \cmark & \cmark & \xmark & \xmark \\
ToolRL\citep{qian2025toolrl} & \cmark & \cmark & \xmark & \xmark \\
\rowcolor{gray!6}
AutoTIR\citep{wei2025autotir} & \cmark & \cmark & \xmark & \xmark \\
SimpleTIR\citep{xue2025simpletir} & \xmark & \cmark & \xmark & \xmark \\
\rowcolor{gray!6}
Agentic Reasoning\citep{wu2025agentic} & \cmark & \cmark & \xmark & \xmark \\
Tool-Star\citep{dong2025tool} & \cmark & \cmark & \xmark & \xmark \\
\rowcolor{gray!6}
Search-R1\citep{jin2025search} & \cmark & \xmark & \xmark & \xmark \\
OTC-PO\citep{wang2025acting} & \cmark & \cmark & \xmark & \xmark \\
\rowcolor{gray!6}
ReasonZoo\citep{zhao2025dissecting} & \cmark & \cmark & \xmark & \xmark \\

\bottomrule
\end{tabular}
\end{table*}

\begin{figure*}[!t]
    \centering
    \includegraphics[width=\textwidth]{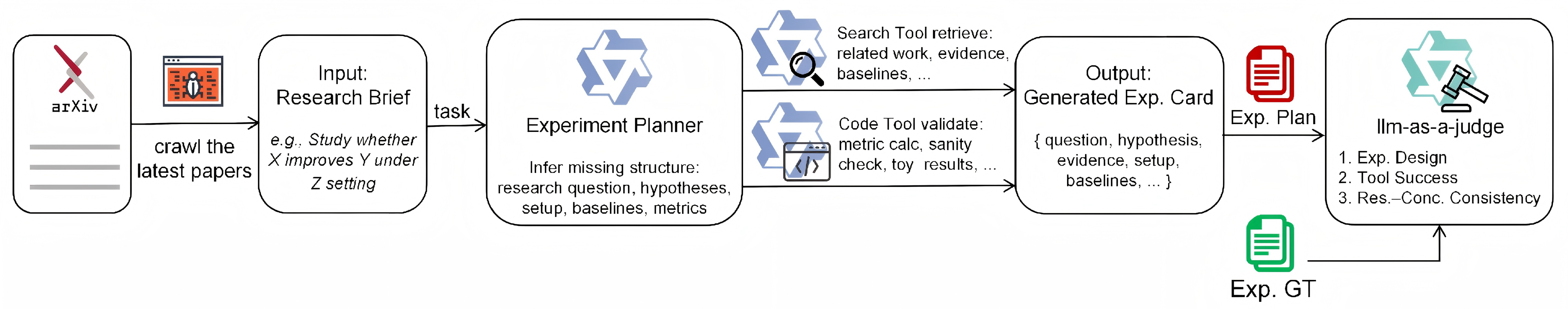}
    \caption{Pipeline of Pipeline of ExpoDesign. Given an underspecified research prompt, the model retrieves external evidence, uses code for lightweight execution or validation, and produces a structured experiment card that is evaluated by an LLM judge for design quality, tool-grounded plausibility, and result--conclusion consistency.}
    \label{fig:open_ended_task}
\end{figure*}

Second, as shown in Table~\ref{tab:benchmarks}, the evaluation of existing methods is largely restricted to QA and mathematical reasoning tasks, where both task complexity and the necessity of tool use remain limited. These benchmarks typically provide only a small set of tools, often search and code execution, and the tasks themselves usually do not require sophisticated tool coordination. For example, as shown in Table~\ref{tab:dataset_diagnostics_updated} in Appendix~\ref{app:full-results}, even when QA tasks allow access to both search and code tools, models tend to rely almost exclusively on the search tool to solve the problems. This suggests that existing TIR tasks are often limited in both task design and tool configuration, making them insufficient for effectively evaluating TIR methods—especially their ability to coordinate multiple tools to support complex reasoning. 
Moreover, compared with real-world decision-making scenarios, traditional math and QA tasks are relatively simple and less representative of practical tool-integrated applications. 
To address these limitations, we introduce two more complex tasks: 

\paragraph{Tool-Grounded Experimental Design (ExpoDesign).}
As illustrated in Figure~\ref{fig:open_ended_task}, this task starts from an underspecified research prompt and asks the system to construct a tool-grounded mini-experiment. For each evaluation case, TIDE-Bench provides a paper-agnostic minimal-experiment brief and an evaluation protocol specifying the interaction format and the expected final deliverable. Given the brief, the system needs to infer a complex structured experiment specification, including a testable research question, falsifiable hypotheses, relevant evidence, data and evaluation metrics, weak and strong baselines, experimental procedure, preliminary results, brief analysis, and limitations.
The central challenge is that the system must not only complete the missing experimental structure from an open-ended research prompt, but also coordinate heterogeneous tools within a single reasoning process. Search is used to ground the research context, collect external evidence, and support hypothesis formation, while code operationalizes assumptions through preliminary checks, metric computation, lightweight simulation, or result validation. During inference, the system may invoke tools over multiple rounds and finally returns a structured experiment card in JSON format.
To prevent retrieval shortcuts, the input omits metadata that could uniquely identify the source paper, such as the title, full abstract, author list, URL, or arXiv ID. We construct the dataset by sampling STEM and computer science papers according to the 2025 arXiv distribution and pairing each record with a human-written or model-generated task brief. Because the task does not admit a unique gold answer, we evaluate each experiment card with an LLM judge along three dimensions: design quality, tool-grounded plausibility, and result--conclusion consistency.

\begin{figure*}[!t]
    \centering
    \includegraphics[width=\textwidth]{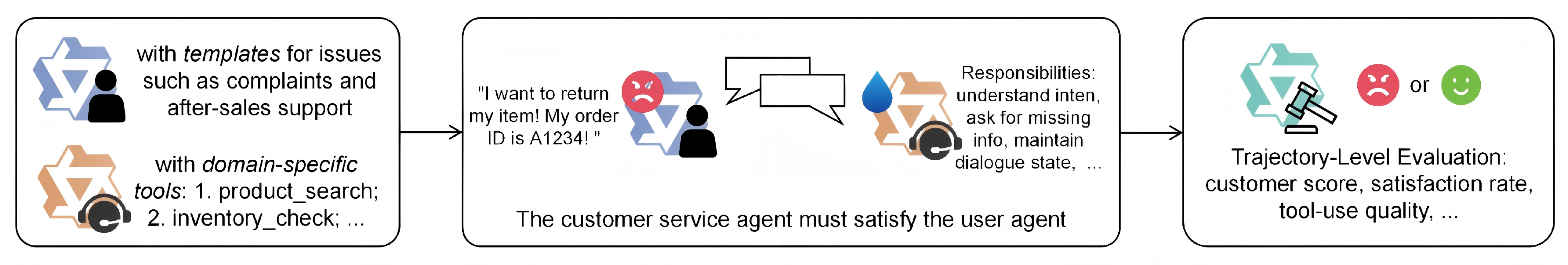}
    \caption{Pipeline of Interaction. In a simulated e-commerce service scenario, the agent interacts with a customer across multiple turns, invokes domain-specific tools to verify products, orders, policies, inventory, and prices, and is evaluated over the full trajectory for dialogue success and tool-use quality.}
    \label{fig:dynamic_interactive_task}
\end{figure*}

\paragraph{Dynamic Interaction (Interaction).}
As illustrated in Figure~\ref{fig:dynamic_interactive_task}, Interaction is defined as a multi-scenario, multi-turn e-commerce service-resolution task. Given a scripted customer scenario, the system acts as a merchant service agent, resolves the user's request through natural-language dialogue, and invokes domain-specific tools when needed to query product, inventory, policy, order, and pricing information. The task covers five service scenarios, including complaints, transactions, after-sales support, clarification, and misunderstanding or adversarial turns, requiring the system to progress toward service resolution over the full dialogue rather than produce a single-turn answer.
In this task, TIR is instantiated as a stateful tool-orchestration problem. The agent must understand evolving user intent, ask for missing information, maintain task state across turns, follow business constraints, and ground user-facing responses in tool outputs. Therefore, the task evaluates whether TIR systems can coordinate multiple tools, track dialogue state, respect domain constraints, and produce tool-supported responses in deployment-oriented interactive settings.
To instantiate this setting, we construct a dialogue-based e-commerce environment in which the evaluated model acts as the merchant service agent, while a fixed API model simulates customers with diverse goals, tones, and information conditions. The benchmark contains 100 scripted scenarios balanced across the five service categories. Each scenario specifies a customer persona, goal, tone, optional missing information, product and order context, and an opening user message.
During each episode, the agent interacts with the customer for up to ten turns and may invoke five user-defined tools: product search, inventory checking, policy retrieval, order lookup, and price calculation. Tool outputs are visible only to the agent and are merged into its context, while the simulated customer observes only the plain dialogue. This design requires the agent to verify facts and constraints through tools before responding, rather than relying on unsupported assumptions.
We evaluate each episode over the full trajectory, including both the dialogue outcome and the tool-use trace. The customer-side score is used as the primary measure of task success, while satisfaction rate and average tool calls serve as auxiliary diagnostics for interaction quality and tool-use behavior.

\subsection{Evaluation Metrics}
\label{sec:eval-open-dynamic}

In this work, we propose a comprehensive and task-aware evaluation protocol that assesses TIR methods from four complementary perspectives: answer quality, process reliability, tool-use efficiency, and computational cost.

\paragraph{Answer quality.}
For mathematical and knowledge-intensive tasks, we adopt an \textbf{LLM-as-a-judge} protocol to determine answer correctness and use the resulting \textbf{accuracy} as the primary evaluation metric. Specifically, the LLM judge produces a binary correctness decision for each example, and the average of these decisions is reported as accuracy. In addition to this primary metric, we retain task-specific diagnostic metrics: exact match (EM), token-level F1, and substring-hit accuracy for knowledge-intensive tasks, and math-level accuracy for mathematical tasks. For ExpoDesign, we use an LLM judge to score \textbf{experimental design}, \textbf{tool success}, and \textbf{result--conclusion consistency} on a 1--5 scale, and use their mean as the overall score. For Interaction, we use an LLM judge to score \textbf{customer-facing success}, \textbf{satisfaction rate}, and \textbf{tool-call behavior} on a 0--5 scale, and use the average of these sub-scores as the \textbf{average customer score}. These decomposed scores allow us to diagnose not only final task success, but also whether models use tools appropriately and produce reliable user-facing responses.


\paragraph{Process reliability.}
Final-answer quality alone is not sufficient for evaluating TIR methods. A model may produce a correct answer while relying on inconsistent reasoning, or invoke tools without actually grounding its response in tool outputs. We therefore introduce two process-level metrics. \textbf{Reasoning--answer consistency} measures whether the final answer is supported by the reasoning trace enclosed in \texttt{<think>...</think>}. \textbf{Tool supportedness} measures whether the final answer is grounded in evidence returned by tools inside \texttt{<result>...</result>}. For QA tasks, tool supportedness allows weak matching through direct containment or sufficiently high token overlap; for mathematical tasks, numeric equivalence is additionally accepted.

\paragraph{Tool-use efficiency.}
Beyond task quality, we explicitly evaluate whether a system uses tools economically. Let $c_i$ denote the number of tool calls in sample $i$, and let $q_i$ denote the task-specific headline quality score: answer correctness for QA and math, the overall judge score for ExpoDesign, and the customer-side dialogue score for Interaction. We define tool efficiency as the average of $q_i/c_i$ over all samples with $c_i > 0$:
\[
\text{ToolEfficiency}
=
\frac{1}{|\{i \mid c_i>0\}|}
\sum_{i:c_i>0}\frac{q_i}{c_i}.
\]
This metric reflects a model's ability to achieve strong task performance while minimizing the number of tool calls.

\paragraph{Computational cost.}
Finally, we report cost-related statistics, since the practical utility of a TIR method depends not only on output quality but also on latency and inference cost. We therefore track the average number of tool calls, total wall-clock time, average latency per sample, and—when tokenization is available—the total and average number of tokens in the prompt and generated trajectory. These metrics support joint analysis of effectiveness and efficiency.

\section{Experiment}

\subsection{Experimental Setup}

\paragraph{Datasets.}
TIDE-Bench establishes a diverse and challenging evaluation environment for TIR systems by including not only cleaned, high-quality mathematical and knowledge-intensive tasks, but also two newly designed tasks that are more complex and better reflect real-world scenarios. The \textbf{mathematical tasks} include GSM8K~\citep{cobbe2021gsm8k}, MATH~\citep{hendrycks2021math}, and Omni-MATH~\citep{gao2024omnimath}. The \textbf{knowledge-intensive tasks} include SimpleQA~\citep{wei2024simpleqa}, HotpotQA~\citep{yang2018hotpotqa}, and SQuAD v2~\citep{rajpurkar2018know}. In addition, we introduce two new task settings, \textbf{ExpoDesign} and \textbf{Interaction}, which extend TIR evaluation to discovery-oriented and deployment-oriented scenarios.

\paragraph{Models and settings.}
We evaluate eight representative systems, including the Qwen3 series models~\citep{yang2025qwen3} (4B, 8B, and 32B), GPT-4o~\citep{hurst2024gpt}, GPT-5~\citep{singh2025openai}, GPT-5.4~\citep{openai2025gpt54systemcard}, SearchR1~\citep{jin2025search}, and Tool-Star~\citep{dong2025tool}. These systems cover commonly used open-source models of different scales, recent closed-source models, and tool-tuned reasoning systems developed for advanced tool-integrated reasoning. For mathematical and knowledge-intensive datasets, each model is evaluated in a matched \emph{no-tool} and \emph{tool-enabled} setting, so the reported gain isolates the effect of allowing external tool calls. For ExpoDesign and Interaction,, we report only the tool-enabled setting, since tool invocation results are part of the task-specific evaluation and the no-tool setting is therefore not applicable. All primary results are reported in Table~\ref{tab:main-results}; detailed results for diagnostic metrics are provided in Appendix~\ref{app:full-results}.

In our experiments, we use \texttt{kimi-k2-0905-preview} as the LLM judge for both answer correctness evaluation and task-specific quality scoring. The main results table reports the primary metric for each task type: accuracy for mathematical and knowledge-intensive tasks, the overall score for ExpoDesign, and the average customer score for Interaction.

\subsection{Main Results}

Table~\ref{tab:main-results} presents the primary evaluation results on TIDE-Bench. The results show that, beyond the relatively high scores on GSM8K and MATH, the reconstructed benchmark contains several challenging tasks where current TIR systems have substantial room for improvement. On Omni-MATH, the best no-tool score is only 0.684, and most open-source or tool-specialized systems obtain much lower accuracy, indicating that harder mathematical reasoning is still far from solved. The knowledge-intensive QA tasks also expose different bottlenecks: SimpleQA and HotpotQA remain difficult for many systems without tool access, whereas SQuAD\_v2 is challenging because it requires reliable no-answer judgment rather than simple factual generation. The two newly introduced tracks are also far from saturated. On ExpoDesign, only GPT-5.4 and GPT-5 achieve scores above 4.5, while all open-source and tool-specialized baselines remain below 3.1. On Interaction, the scores range widely from 1.620 to 4.530, showing large differences in interactive tool-use performance. Overall, these results indicate that TIDE-Bench contains multiple challenging tasks that current TIR systems cannot yet solve reliably, highlighting the need for further research on tool-grounded reasoning, evidence integration, and interactive tool use.

\begin{figure*}[t]
    \centering
    \includegraphics[width=0.98\textwidth]{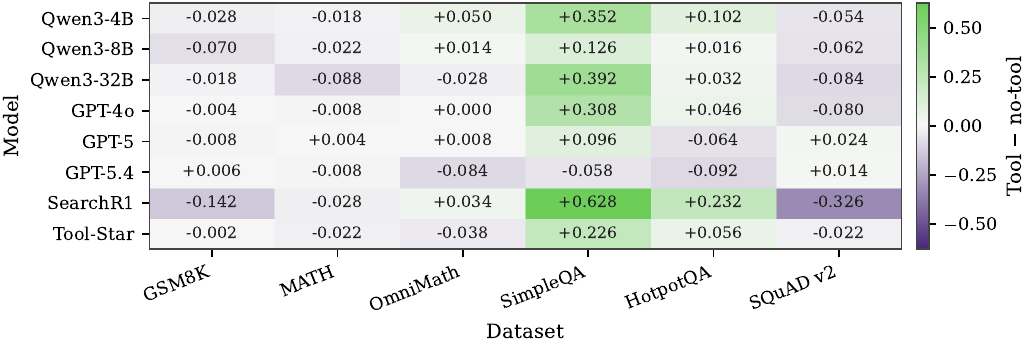}
    \caption{Tool gain over the no-tool baseline on mathematical and knowledge-intensive tasks. Each cell shows the difference between the tool-enabled and no-tool primary metric for a given model--dataset pair. Positive values indicate performance gains from tool access, while negative values indicate degradation.}
    \label{fig:tool_gain_heatmap}
\end{figure*}

By comparing the matched tool-enabled and no-tool results, we find that tool access does not lead to universal improvement; its effect depends on whether the available tool matches the main task bottleneck. As shown in Table~\ref{tab:main-results} and Figure~\ref{fig:tool_gain_heatmap}, the most consistent improvement appears on SimpleQA, where enabling tools raises the average accuracy by $+0.259$ across systems. SearchR1, Qwen3-32B, Qwen3-4B, and GPT-4o all obtain large gains on this task, indicating that the search tool is well aligned with short-form factual QA and that such tasks strongly depend on external evidence. HotpotQA also improves on average, but the gain is much smaller ($+0.041$), suggesting that retrieval alone becomes less sufficient when the model must integrate evidence across multiple hops. In contrast, the gains on mathematical reasoning are weak and unstable: GSM8K and MATH slightly degrade on average, while Omni-MATH changes only marginally. This suggests that code execution does not automatically improve mathematical reasoning unless the model can first formalize the problem into a correct executable procedure. SQuAD v2 also drops by $-0.074$ on average, because additional retrieved passages may interfere with the model's no-answer judgment rather than provide useful evidence. These results suggest that future TIR systems should not only improve tool access, but also enhance tool design, tool-grounding ability, and the ability to determine when tool use is necessary or beneficial.

\begin{table*}[t]
\centering
\caption{
Main results on TIDE-Bench across eight datasets.Mathematical and knowledge-intensive datasets are evaluated by accuracy, while the two newly designed datasets use task-specific scores. SearchR1 and Tool-Star are TIR-oriented fine-tuned baselines, specializing in search-based and multi-tool reasoning, respectively.
}
\label{tab:main-results}
\small
\resizebox{\textwidth}{!}{%
\begin{tabular}{llcccccccc}
\toprule
\multirow{3}{*}{Setting} & \multirow{3}{*}{Model} & \multicolumn{3}{c}{Math} & \multicolumn{3}{c}{Knowledge-intensive QA} & \multicolumn{2}{c}{Ours} \\ \cmidrule(lr){3-5}\cmidrule(lr){6-8}\cmidrule(lr){9-10}
 & & GSM8K & MATH & Omni & SimpleQA & HotpotQA & SQuAD\_v2 & ExpoDesign & Interaction \\ 
 & & Acc. & Acc. & Acc. & Acc. & Acc. & Acc. & Score & Score \\ \midrule
\multirow{8}{*}{No-Tool} & Qwen3-4B & 0.896 & 0.938 & 0.308 & 0.038 & 0.226 & 0.702 & \na & \na \\
 & Qwen3-8B & 0.920 & 0.936 & 0.322 & 0.052 & 0.242 & 0.732 & \na & \na \\
 & Qwen3-32B & 0.892 & 0.968 & 0.348 & 0.058 & 0.254 & 0.736 & \na & \na \\
 & GPT-4o & 0.926 & 0.774 & 0.174 & 0.424 & 0.494 & 0.770 & \na & \na \\
 & GPT-5 & 0.948 & 0.984 & 0.638 & 0.510 & 0.566 & 0.714 & \na & \na \\
 & GPT-5.4 & 0.942 & 0.986 & 0.684 & 0.462 & 0.554 & 0.692 & \na & \na \\
 & SearchR1 & 0.476 & 0.380 & 0.114 & 0.080 & 0.208 & 0.546 & \na & \na \\
 & Tool-Star & 0.852 & 0.754 & 0.190 & 0.434 & 0.328 & 0.618 & \na & \na \\ \hline
\multirow{8}{*}{Tool} & Qwen3-4B & 0.868 & 0.920 & 0.358 & 0.390 & 0.328 & 0.648 & 2.831 & 3.790 \\
 & Qwen3-8B & 0.850 & 0.914 & 0.336 & 0.178 & 0.258 & 0.670 & 2.984 & 4.080 \\
 & Qwen3-32B & 0.874 & 0.880 & 0.320 & 0.450 & 0.286 & 0.652 & 3.032 & 4.530 \\
 & GPT-4o & 0.922 & 0.766 & 0.174 & 0.732 & 0.540 & 0.690 & 2.523 & 1.620 \\
 & GPT-5 & 0.940 & 0.988 & 0.646 & 0.606 & 0.502 & 0.738 & 4.680 & 2.770 \\
 & GPT-5.4 & 0.948 & 0.978 & 0.600 & 0.404 & 0.462 & 0.706 & 4.785 & 3.570 \\
 & SearchR1 & 0.334 & 0.352 & 0.148 & 0.708 & 0.440 & 0.220 & 1.773 & 2.100 \\
 & Tool-Star & 0.850 & 0.732 & 0.152 & 0.660 & 0.384 & 0.596 & 2.531 & 2.460 \\ \hline
\end{tabular}
}
\end{table*}

The two newly introduced task tracks produce rankings that differ from standard math and QA datasets, indicating that they reveal task-specific bottlenecks beyond closed-ended evaluation. On ExpoDesign, GPT-5.4 and GPT-5 achieve the highest scores, 4.785 and 4.680, while all open-source and tool-specialized baselines remain below 3.1. This gap reflects the open-ended nature of ExpoDesign, where models must plan experiments, retrieve and synthesize evidence, perform lightweight validation with code, and align conclusions with intermediate results. In contrast, the Interaction is led by Qwen3-32B, which obtains the highest customer-facing score of 4.530, outperforming GPT-series and tool-specialized models. This suggests that deployment-oriented interaction stresses different abilities, including tracking evolving user intent, deciding when tools are needed, maintaining state across turns, following business constraints, and grounding responses in tool outputs. Together, these results show that TIR ability is multi-dimensional: retrieval QA, mathematical reasoning, experimental design, and interactive service resolution each test different aspects of tool-integrated reasoning.


\subsection{Benefit--Cost Trade-offs of Tool Use}

\begin{figure*}[t]
    \centering
    \includegraphics[width=0.98\textwidth]{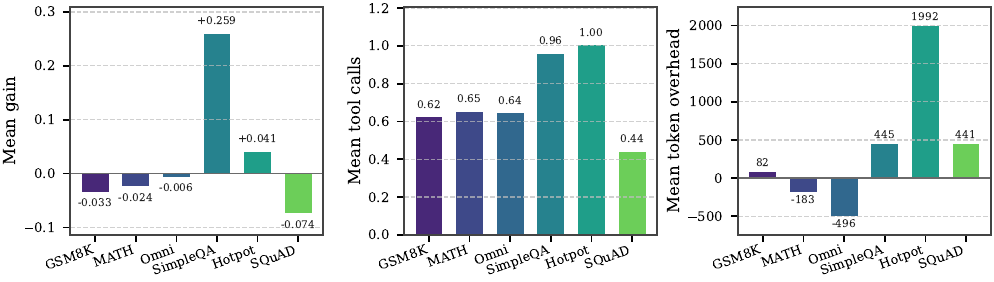}
    \caption{Performance gains and computational costs of tool use across mathematical and knowledge-intensive datasets. 
    Left: average performance gain from enabling tool use relative to the no-tool setting. 
    Middle: mean number of tool calls in the tool-enabled setting. 
    Right: mean change in token usage introduced by tool use, where negative values indicate token savings.}
    \label{fig:three_panel_tradeoff}
\end{figure*}



Although tool invocation can improve task performance, it may
also introduce additional computational costs, including more tool calls, longer contexts, and larger token overhead. This motivates our benefit--cost trade-off analysis of tool use, rather than evaluating tool use solely by performance gain. As shown in Figure~\ref{fig:three_panel_tradeoff}, SimpleQA achieves the most significant average gain with moderate cost, whereas HotpotQA incurs the highest token overhead but obtains only a relatively limited performance improvement. Both tasks involve the search tool, but their returns differ: SimpleQA can often be answered once the model retrieves a single decisive fact, while HotpotQA still requires multi-hop filtering and evidence composition after retrieval.

The negative or weak gains on mathematical tasks and SQuAD v2 highlight the importance of deciding \emph{when tools are actually necessary}. For mathematical tasks, tool calls may reduce the length and token cost of the reasoning process without necessarily improving the final answer. This helps explain why the token overhead for MATH and Omni-MATH can be negative in Figure~\ref{fig:three_panel_tradeoff}: Python execution may replace part of the model's long-form textual reasoning with shorter code-and-result traces, especially on harder problems. However, the weak performance gains indicate that lower generation cost does not necessarily translate into better reasoning quality. For SQuAD v2, additional retrieval can be actively harmful because the model must determine whether a question is answerable at all. These cases suggest that future TIR training should optimize not only tool-call success, but also call necessity, evidence selection, and abstention under insufficient support.

In addition to the metrics discussed above, ExpoDesign and Interaction also provide a range of fine-grained diagnostic metrics, which are introduced in the \textit{Answer Quality} subsection of our methodology. These metrics enable a more comprehensive evaluation of the TIR reasoning process, and the detailed results are provided in Appendix~\ref{app:full-results}.

\section{Conclusion}
\label{sec:conclusion}
We presented TIDE-Bench, a task-aware and diagnostic benchmark for Tool-Integrated Reasoning. It combines compact mathematical reasoning and knowledge-intensive QA subsets with two tool-mediated tracks, ExpoDesign and Interaction, and evaluates final quality, process reliability, tool-use efficiency, and inference cost under a unified protocol.
Our experiments show that tool use is beneficial only when it matches the task bottleneck. Search helps most on direct factual QA, but is less reliable for multi-hop evidence composition and answerability-sensitive QA; code execution helps mathematical reasoning only when the problem is first formalized correctly. The two new tracks further expose bottlenecks in evidence-grounded experimental planning and stateful multi-turn tool orchestration.
These findings suggest that future TIR systems should optimize selective, grounded, and cost-aware tool use rather than tool-call frequency alone. TIDE-Bench provides an efficient testbed for diagnosing when tools improve reasoning, when they fail, and which bottlenecks future methods should address.

\clearpage

\bibliographystyle{plainnat}
\bibliography{custom}
\newpage
\appendix
\section{Implement Details}
\label{sec:appendix}
\paragraph{Implement Details in Data Cleaning.} During data cleaning, we used the Qwen3-4B model to evaluate the Math reasoning tasks and knowledge-intensive tasks, with the parameters set to temperature = 0.0, top p = 1.0, and max tokens = 8192. For the datasets in both categories, we sorted the entries in ascending order based on their F1 scores and accuracy, respectively, and selected the first 500 entries as the final cleaned dataset. Throughout this process, the model did not possess tool-calling capabilities.
\paragraph{Implement Details in Experiments.} In the experimental section, when use tool is set to false, the model parameters are configured as follows: temperature = 0.2, max tokens = 2048, top p = 0.95, and top k = 20. When use tool is set to true, models are permitted to solve problems by invoking tools via system prompts. In this configuration, the max tokens limit is adjusted to 3072 for Qwen3-8B, and to 4096 for all other open-source and closed-source models. All other settings remain unchanged.

During tool invocation, the maximum number of calls for the code tool is set to 5, while the limit for all other tools is set to 3. For knowledge-intensive tasks involving local search, we employ E5-base-v2 \citep{wang2022text} with Wikipedia retrieval following the FlashRAG configuration \citep{jin2025flashrag}. For all other tasks, Bing Web Search is utilized for retrieval.

The hyperparameters used in the data cleaning stage differ from those in the main experiments. This is because the cleaning process relies on a fixed configuration with Qwen3-4B as a reference model to identify low-complexity instances, which may not generalize well to other models. For the main evaluation, we instead adopt a unified set of hyperparameters that is consistently applied across all evaluated models, ensuring a fair and stable comparison.

\section{Tools under the Tag-based Protocol}
\label{app:user_defined_tools}

To ensure broad compatibility across models with different tool-use capabilities, we adopt a lightweight tag-based tool invocation protocol following the implementation of ARPO~\citep{dong2025arpo}. While many contemporary closed-source models provide native function-calling mechanisms via structured JSON formats~\citep{wang2025function}, such interfaces are not consistently available in open-source or locally deployed models, especially base model families without native function-calling support~\citep{adcock2026llama,yang2025qwen3}. Therefore, instead of relying on model-specific function-calling APIs, we require all models to invoke tools using a unified XML-like tag format:
\begin{equation}
\texttt{<tool\_name>\{arguments\}</tool\_name>}.
\end{equation}

In this protocol, the enclosed content can be either JSON-formatted arguments or a plain-text query. If the content is valid JSON, it is parsed as a dictionary; otherwise, it is treated as a query string. Our orchestration layer detects the predefined tags, parses the arguments, executes the corresponding tool, and wraps the returned result within \texttt{<result></result>} tags. This design provides a model-agnostic interface for tool usage, enabling different models to interact with external tools under the same protocol. Based on this protocol, we implement two general-purpose tools, namely a code execution tool and a search tool. Examples of invoking these two tools are provided in Appendix~\ref{CaseStudy}.

Beyond these general-purpose tools, we further instantiate the same protocol with five user-defined tools for Interaction in a synthetic e-commerce environment. These tools simulate common service capabilities, including product search, inventory checking, policy retrieval, order lookup, and price calculation. They operate over shared synthetic resources, including a product catalog, store policies, and coupon rules. The product catalog covers several common domains, such as electronics, home and kitchen, beauty and personal care, fashion, sports and outdoors, pet supplies, and office supplies.

\begin{itemize}
    \item \textbf{\texttt{product\_search}} retrieves relevant products from the synthetic catalog according to a user query, with an optional domain filter and a maximum number of returned items.

    \item \textbf{\texttt{inventory\_check}} simulates stock availability for a product based on its SKU or product name, returning the stock status, available quantity, and estimated restock time if unavailable.

    \item \textbf{\texttt{policy\_search}} retrieves the most relevant store policy according to a user-provided topic or query, covering returns, shipping, warranty, price matching, and coupons.

    \item \textbf{\texttt{order\_lookup}} simulates order information retrieval for a given order ID, including order status, purchased item, purchase time, tracking number, and whether the shipping address can still be changed.

    \item \textbf{\texttt{pricing\_calc}} computes the final order total based on product price, quantity, coupon code, and shipping fee, supporting percentage discounts, fixed discounts, and free-shipping coupons.
\end{itemize}

Table~\ref{tab:user_defined_tools} summarizes the five user-defined tools.

\begin{table}[h]
\centering
\caption{Overview of the user-defined tools used in Interaction.}
\small
\begin{tabular}{lll}
\hline
\textbf{Tool} & \textbf{Main Function} & \textbf{Typical Input} \\
\hline
\texttt{product\_search} & Search products in the catalog & query, domain, k \\
\texttt{inventory\_check} & Check stock availability & sku, product\_name \\
\texttt{policy\_search} & Retrieve store policies & topic or query \\
\texttt{order\_lookup} & Look up order status & order\_id \\
\texttt{pricing\_calc} & Calculate price, discount, and shipping & sku, quantity, coupon\_code \\
\hline
\end{tabular}
\label{tab:user_defined_tools}
\end{table}

\section{Full Experimental Results}
\label{app:full-results}

This appendix reports the score-level results and diagnostics from the updated tidy result table. Tables are preferred over additional figures because the appendix is meant to make numerical values directly checkable, while the main paper uses figures only for aggregate trends and trade-offs. The complete machine-readable result file is included in the source package as \texttt{results/tidy\_results.csv}.

\begin{table*}[h]
\centering
\small
\caption{Tool-enabled gain over the matched no-tool baseline on mathematical and knowledge-intensive datasets. Each cell is tool minus no-tool headline accuracy; the final column averages the six gains for each model.}
\label{tab:closed_gain_updated}
\begin{tabular}{lrrrrrrr}
\toprule
Model & GSM8K & MATH & Omni & SimpleQA & HotpotQA & SQuAD v2 & Avg. gain \\
\midrule
Qwen3-4B & -0.028 & -0.018 & +0.050 & +0.352 & +0.102 & -0.054 & +0.067 \\
Qwen3-8B & -0.070 & -0.022 & +0.014 & +0.126 & +0.016 & -0.062 & +0.000 \\
Qwen3-32B & -0.018 & -0.088 & -0.028 & +0.392 & +0.032 & -0.084 & +0.034 \\
GPT-4o & -0.004 & -0.008 & +0.000 & +0.308 & +0.046 & -0.080 & +0.044 \\
GPT-5 & -0.008 & +0.004 & +0.008 & +0.096 & -0.064 & +0.024 & +0.010 \\
GPT-5.4 & +0.006 & -0.008 & -0.084 & -0.058 & -0.092 & +0.014 & -0.037 \\
SearchR1 & -0.142 & -0.028 & +0.034 & +0.628 & +0.232 & -0.326 & +0.066 \\
Tool-Star & -0.002 & -0.022 & -0.038 & +0.226 & +0.056 & -0.022 & +0.033 \\
\bottomrule
\end{tabular}
\end{table*}

\begin{table*}[h]
\centering
\small
\caption{Dataset-level process and cost diagnostics for tool-enabled mathematical and knowledge-intensive tasks, averaged over the eight evaluated methods. Avg. gain is computed against matched no-tool baselines.}
\label{tab:dataset_diagnostics_updated}
\resizebox{\textwidth}{!}{%
\begin{tabular}{lrrrrrrrr}
\toprule
Dataset & Avg. gain & Tool calls & Python calls & Search calls & Supported & Efficiency & Time (s) & Tokens \\
\midrule
GSM8K & -0.033 & 0.624 & 0.622 & 0.002 & 0.515 & 0.559 & 25.80 & 1078.99 \\
MATH & -0.024 & 0.652 & 0.646 & 0.006 & 0.432 & 0.708 & 77.63 & 1894.33 \\
Omni & -0.006 & 0.644 & 0.639 & 0.005 & 0.257 & 0.176 & 252.16 & 6162.42 \\
SimpleQA & +0.259 & 0.959 & 0.021 & 0.938 & 0.639 & 0.556 & 100.11 & 1199.48 \\
HotpotQA & +0.041 & 1.004 & 0.010 & 0.994 & 0.656 & 0.258 & 141.45 & 2685.60 \\
SQuAD v2 & -0.074 & 0.441 & 0.028 & 0.413 & 0.309 & 0.047 & 37.47 & 1199.41 \\
\bottomrule
\end{tabular}%
}
\end{table*}

\begin{table*}
\centering
\small
\caption{QA diagnostic metrics on SimpleQA, HotpotQA, and SQuAD v2. F1 and EM are string-based diagnostic metrics, while LLM Acc. reports the LLM-as-judge binary accuracy used as the headline QA metric. All metrics are reported on the 0--1 scale.}
\label{tab:qa_diagnostics_updated}
\resizebox{\textwidth}{!}{%
\begin{tabular}{llccccccccc}
\toprule
Setting & Model & \multicolumn{3}{c}{SimpleQA} & \multicolumn{3}{c}{HotpotQA} & \multicolumn{3}{c}{SQuAD v2} \\
\cmidrule(lr){3-5}\cmidrule(lr){6-8}\cmidrule(lr){9-11}
 & & F1 & EM & LLM Acc. & F1 & EM & LLM Acc. & F1 & EM & LLM Acc. \\
\midrule
\multirow{8}{*}{no-tool} & Qwen3-4B & 0.079 & 0.020 & 0.038 & 0.166 & 0.100 & 0.226 & 0.105 & 0.078 & 0.702 \\
 & Qwen3-8B & 0.096 & 0.040 & 0.052 & 0.197 & 0.130 & 0.242 & 0.116 & 0.096 & 0.732 \\
 & Qwen3-32B & 0.091 & 0.032 & 0.058 & 0.191 & 0.122 & 0.254 & 0.112 & 0.094 & 0.736 \\
 & GPT-4o & 0.444 & 0.304 & 0.424 & 0.373 & 0.254 & 0.494 & 0.106 & 0.072 & 0.770 \\
 & GPT-5 & 0.468 & 0.300 & 0.510 & 0.399 & 0.258 & 0.566 & 0.106 & 0.072 & 0.714 \\
 & GPT-5.4 & 0.437 & 0.288 & 0.462 & 0.383 & 0.246 & 0.554 & 0.110 & 0.072 & 0.692 \\
 & SearchR1 & 0.110 & 0.056 & 0.080 & 0.154 & 0.090 & 0.208 & 0.112 & 0.084 & 0.546 \\
 & Tool-Star & 0.424 & 0.294 & 0.434 & 0.266 & 0.192 & 0.328 & 0.102 & 0.078 & 0.618 \\
\midrule
\multirow{8}{*}{tool} & Qwen3-4B & 0.362 & 0.258 & 0.390 & 0.230 & 0.164 & 0.328 & 0.088 & 0.064 & 0.648 \\
 & Qwen3-8B & 0.175 & 0.118 & 0.178 & 0.185 & 0.130 & 0.258 & 0.103 & 0.076 & 0.670 \\
 & Qwen3-32B & 0.398 & 0.274 & 0.450 & 0.203 & 0.148 & 0.286 & 0.097 & 0.076 & 0.652 \\
 & GPT-4o & 0.684 & 0.500 & 0.732 & 0.373 & 0.246 & 0.540 & 0.099 & 0.062 & 0.690 \\
 & GPT-5 & 0.537 & 0.382 & 0.606 & 0.310 & 0.210 & 0.502 & 0.116 & 0.088 & 0.738 \\
 & GPT-5.4 & 0.377 & 0.250 & 0.404 & 0.350 & 0.242 & 0.462 & 0.118 & 0.078 & 0.706 \\
 & SearchR1 & 0.667 & 0.520 & 0.708 & 0.345 & 0.252 & 0.440 & 0.054 & 0.046 & 0.220 \\
 & Tool-Star & 0.616 & 0.474 & 0.660 & 0.308 & 0.232 & 0.384 & 0.089 & 0.062 & 0.596 \\
\bottomrule
\end{tabular}%
}
\end{table*}

\begin{table*}[t]
\centering
\small
\caption{Model-level process and cost diagnostics for tool-enabled mathematical and knowledge-intensive tasks, averaged over the six datasets. Avg. gain is computed against matched no-tool baselines.}
\label{tab:model_diagnostics_updated}
\resizebox{\textwidth}{!}{%
\begin{tabular}{lrrrrrrrr}
\toprule
Model & Avg. gain & Tool calls & Python calls & Search calls & Supported & Efficiency & Time (s) & Tokens \\
\midrule
Qwen3-4B & +0.067 & 0.385 & 0.013 & 0.371 & 0.645 & 0.404 & 193.48 & 3956.66 \\
Qwen3-8B & +0.000 & 0.222 & 0.024 & 0.198 & 0.662 & 0.532 & 121.15 & 3571.04 \\
Qwen3-32B & +0.034 & 0.537 & 0.144 & 0.393 & 0.545 & 0.426 & 214.79 & 3437.01 \\
GPT-4o & +0.044 & 1.029 & 0.489 & 0.539 & 0.322 & 0.430 & 38.08 & 1674.77 \\
GPT-5 & +0.010 & 0.312 & 0.085 & 0.227 & 0.355 & 0.331 & 54.05 & 807.76 \\
GPT-5.4 & -0.037 & 0.012 & 0.005 & 0.008 & 0.314 & 0.369 & 27.73 & 505.27 \\
SearchR1 & +0.066 & 2.112 & 1.274 & 0.838 & 0.377 & 0.186 & 64.79 & 1529.38 \\
Tool-Star & +0.033 & 1.154 & 0.586 & 0.568 & 0.564 & 0.395 & 132.10 & -- \\
\bottomrule
\end{tabular}%
}
\end{table*}

\clearpage

\section{Case Study}
\label{CaseStudy}

To provide a more detailed illustration of the tools used in our framework, we present two representative examples drawn from the reasoning outputs of Qwen3-4B on the MATH500 and SimpleQA datasets, respectively. Tables~\ref{tab:case-math} and~\ref{tab:case-search} demonstrate how the model leverages the code tool and the search tool to solve problems.

\begin{table*}[h]
\centering
\caption{An example of Qwen3-4B reasoning on the MATH500 dataset with Python as the tool.}
\label{tab:case-math}

\begin{minipage}{0.95\textwidth}
\setlength{\parindent}{0pt}

\hrule
\vspace{0.4em}

\textbf{Example \#1 from MATH500 Dataset}

\vspace{0.4em}
\hrule
\vspace{0.4em}

\textbf{Question:}

The square root of $t$ is greater than 2 and less than 3.5. How many integer values of $t$ satisfy this condition?

\textbf{Golden Answer:}

8

\vspace{0.4em}
\hrule
\vspace{0.4em}

\textbf{Qwen3-4b's Strategy}

\vspace{0.4em}
\hrule
\vspace{0.6em}

\textless{}think\textgreater{}

To determine how many integer values of \(t\) satisfy the condition \(2 < \sqrt{t} < 3.5\), we can follow these steps:

1. Square all parts of the inequality to eliminate the square root:
\[
2^2 < t < (3.5)^2
\]

This simplifies to:
\[
4 < t < 12.25
\]

2. Identify the integer values of \(t\) that fall within this range. The integers between 4 and 12.25 are 5, 6, 7, 8, 9, 10, 11, and 12.

3. Count the number of these integers.

Let's implement this in Python to ensure accuracy.

\textless{}/think\textgreater{}\textless{}python\textgreater{}

\begin{pycode}
# Define the range
lower_bound = 4
upper_bound = 12.25

# Find the integer values in the range
integer_values = [t for t in range(int(lower_bound) + 1, int(upper_bound) + 1)]

# Count the number of integer values
num_integer_values = len(integer_values)
print(num_integer_values)
\end{pycode}
\textless{}/python\textgreater{}\textless{}result\textgreater{}8\textless{}/result\textgreater{}\textless{}answer\textgreater{}The number of integer values of \(t\) that satisfy the condition \(2 < \sqrt{t} < 3.5\) is \(\boxed{8}\).\textless{}/answer\textgreater{}

\vspace{0.3em}
\hrule
\end{minipage}
\end{table*}

\begin{table*}[t]
\centering
\caption{An example of Qwen3-4B using the search tool for reasoning.}
\label{tab:case-search}

\begin{minipage}{0.95\textwidth}
\setlength{\parindent}{0pt}

\hrule
\vspace{0.4em}

\textbf{Example \#1 from SimpleQA dataset}

\vspace{0.4em}
\hrule
\vspace{0.4em}

\textbf{Question:}

When did Sara Duterte resign from Lakas-CMD?

\textbf{Golden Answer:}

May 19, 2023

\vspace{0.4em}
\hrule
\vspace{0.4em}

\textbf{Qwen3-4B's Strategy}

\vspace{0.4em}
\hrule
\vspace{0.6em}

\textless{}think\textgreater{}

Okay, I need to find out the date when Sara Duterte resigned from Lakas-CMD. First, I should check if there's any public information or news articles about her resignation. Let me start by searching for relevant information.

I'll use the Wikipedia search tool to look up ``Sara Duterte'' and see if there's any mention of her resignation.

\textless{}/think\textgreater{}\textless{}search\textgreater{}Sara Duterte resignation Lakas-CMD\textless{}/search\textgreater{}

\textless{}result\textgreater{}Page 1: Jun 12, 2023 · On May 19, Vice President and Education Secretary Sara Duterte announced her irrevocable resignation from Lakas-Christian Muslim Democrats (Lakas-CMD), the political party \ldots{} Page 2: May 20, 2023 · Following a shakeup that saw her close ally, former president Gloria Macapagal-Arroyo, lose a senior House leadership position, Vice President Sara Duterte on Friday, May 19, resigned \ldots{}\textless{}/result\textgreater{}

\textless{}think\textgreater{}

From the search results, it seems that Sara Duterte resigned from Lakas-CMD on May 19, 2023. The first result mentions that she announced her resignation on May 19, 2023. So the day is May 19th, the month is May, and the year is 2023. I don't think there's any conflicting information here. The second result also confirms the same date. Therefore, the answer should be May 19, 2023.

\textless{}/think\textgreater{}\textless{}answer\textgreater{}The final answer is \(\boxed{\text{May 19, 2023}}\).\textless{}/answer\textgreater{}

\vspace{0.3em}
\hrule
\end{minipage}
\end{table*}

\end{document}